\RequirePackage{amsmath}
\documentclass[runningheads]{llncs}
\usepackage[T1]{fontenc}
\usepackage{tabularx}
\usepackage{array, boldline, booktabs, makecell}
\usepackage{multirow}
\usepackage{amsmath,amssymb,amsfonts}
\usepackage{graphicx}
\usepackage{textcomp}
\usepackage{xcolor}
\usepackage{hyperref, etoolbox}
\def\BibTeX{{\rm B\kern-.05em{\sc i\kern-.025em b}\kern-.08em
    T\kern-.1667em\lower.7ex\hbox{E}\kern-.125emX}}

\usepackage{newfloat}
\usepackage{listings}

\usepackage{booktabs}
\usepackage{algorithm}
\usepackage{algorithmic}
\usepackage{multirow}

\usepackage{mathrsfs}

\usepackage{tabularx, ragged2e}
\usepackage{verbatim}
\usepackage{fancyvrb}
\usepackage{fvextra}

\usepackage{amsmath,amsfonts,bm}

\def\eqref#1{(\ref{#1})}
\def\Eqref#1{Equation~(\ref{#1})}
\def\1{\bm{1}}

\DeclareMathAlphabet{\mathsfit}{\encodingdefault}{\sfdefault}{m}{sl}
\SetMathAlphabet{\mathsfit}{bold}{\encodingdefault}{\sfdefault}{bx}{n}

\newcommand{\E}{\mathbb{E}}

\begin{document}

\title{Unlearnable Examples For Time Series}

\author{{Yujing Jiang\inst{1}, Xingjun Ma\inst{2}, Sarah Monazam Erfani\inst{1}, James Bailey\inst{1}}}
\authorrunning{Y. Jiang et al.}
\institute{\textsuperscript{1} Faculty of Engineering and Information Technology, The University of Melbourne \\
\email{\{yujingj@student., sarah.erfani@, baileyj@\}unimelb.edu.au} \\ \textsuperscript{2} School of Computer Science, Fudan University \\ \email{xingjunma@fudan.edu.cn}}

\maketitle

\begin{abstract}

Unlearnable examples (UEs) refer to training samples modified to be unlearnable to Deep Neural Networks (DNNs). These examples are usually generated by adding error-minimizing noises that can fool a DNN model into believing that there is nothing (no error) to learn from the data.
The concept of UE has been proposed as a countermeasure against unauthorized data exploitation on personal data.
While UE has been extensively studied on images, it is unclear how to craft effective UEs for time series data.
In this work, we introduce the first UE generation method to protect time series data from unauthorized training by deep learning models. To this end, we propose a new form of error-minimizing noise that can be \emph{selectively} applied to specific segments of time series, rendering them unlearnable to DNN models while remaining imperceptible to human observers. Through extensive experiments on a wide range of time series datasets, we demonstrate that the proposed UE generation method is effective in both classification and generation tasks. It can protect time series data against unauthorized exploitation, while preserving their utility for legitimate usage, thereby contributing to the development of secure and trustworthy machine learning systems.

\end{abstract}

\section{Introduction}

The rapid advancement of deep learning and large models is largely driven by the vast amounts of data ``freely'' available on the Internet. While there has been significant research aimed at training deep learning models with privacy preservation~\cite{shokri2015privacy,abadi2016deep,phan2016differential,phan2017adaptive,shokri2017membership}, these approaches still neglect the necessity to obtain users' consent to use their data. Recent works have proposed useful tools such as Fawkes \cite{shan2020fawkes} to address this gap by promoting consent-based data utilization and protection. Yet, the issue remains unresolved. Rising public concerns stem from several instances where personal data, harvested from the Internet without consent, has been utilized to train commercial machine learning models~\cite{hill_2020}. Concerns now encompass not only images but also time series and multi-modal data. This broadening scope underscores the need for thorough data protection strategies, particularly in the relatively underexplored field of time series data.

%Existing research has primarily focused on image-based applications due to the vital role of image data in fields such as face recognition \cite{balaban2015deep,trigueros2018face}, surveillance \cite{nawaratne2019spatiotemporal,xu2021deep}, and health diagnostics \cite{kaul2022deep,liu2022deep}.
In response to privacy concerns, a number of data protection techniques have been developed including secure release and protective data poisoning. Among those works, protective data poisoning techniques have become increasingly attractive as they allow users to actively add poisoning or adversarial noise into their data (like selfies) before posting them on online social media platforms to protect data exploits. Recently, more advanced data protection techniques such as Unlearnable Examples (UEs)~\cite{huang2020unlearnable,ren2022transferable,fu2022robust,zhang2023unlearnable} have been proposed which can make (image) data unlearnable to machine learning models. Contrasting this with conventional data protection techniques that simply obscure identifiable data, UEs ensure that a DNN trained on such examples performs no better than random guessing on standard test examples.

Existing research on either data poisoning-based data protection or UEs has primarily focused on image-based applications, overlooking the significance of time series data which is vital in applications such as financial forecasting \cite{barra2020deep}, health monitoring \cite{maweu2021generating}, energy prediction \cite{feng2021space}, and transportation \cite{ma2020hybrid}. Given its distinct characteristics and broad applications, there is an urgent need for time series data protection methods. Image-oriented data protection methods might not translate well to time series data due to their dynamic and sequential nature~\cite{jiang2022backdoor}.
Although the concept of UEs has predominantly been confined to computer vision, our research will demonstrate that this concept can also be effectively extended to time series applications.

Existing methods developed for image-based UEs often apply unlearnable noise across the whole image. However, this approach is less suitable for time series data, which is inherently sequential and often requires interventions in certain segments instead of the entire dataset. Given that a short segment in time series data can hold critical information about a particular process or entity, the direct application of image-based UE techniques to time series data encounters significant challenges. Recognizing these limitations, we propose to make only a fraction, i.e., the most sensitive or crucial part, of the time series data unlearnable. This allows for the protection of specific data segments while maintaining the usability of the remainder, balancing security with data integrity.

In this work, we extend the concept of UEs to time series data and propose a novel and effective UE generation method. Our contributions are as follows:

\begin{itemize}

\item We introduce a new form of error-minimizing noise that can be applied \emph{selectively} to segments of time series. This noise is imperceptible to humans, preserving the overall utility of the data while ensuring its primary purpose of rendering the data unlearnable to DNNs.

\item We propose a novel unlearnable noise generator that can mitigate the potential risk of the underlying time series data being recognized or trained by either classification or generative models. By applying this noise, we effectively create a layer of protection around the data, making it ineffective for exploitation by AI technologies, while preserving its value for legitimate use.

\item We conduct empirical studies to demonstrate the effectiveness of our method in generating unlearnable examples. Our evaluation covers a broad range of time series datasets, showcasing its versatility and robustness.

\end{itemize}

\section{Related Work}

%The growing concerns regarding ethical data usage have prompted researchers to devise innovative strategies for data privacy.
In this section, we briefly review the most relevant data protection methods including data poisoning, adversarial attacks, and unlearnable examples.

\vspace{-1ex}
\subsection{Data Poisoning}

Data poisoning attacks aim to weaken a model's performance by altering training data. Such attacks on Support Vector Machines (SVM) were shown by \cite{biggio2012poisoning}. Koh et al. \cite{koh2017understanding} expanded this, targeting influential training samples in DNNs with adversarial noise. This was later adopted into an end-to-end framework \cite{munoz2017towards}. The work ``Poison Frogs!'' presents a clean-label poisoning technique that retains correct labels, making the attack more insidious \cite{shafahi2018poison}. Backdoor attacks, another variant of data poisoning techniques, involve embedding a hidden trigger pattern into the training dataset. Despite this manipulation, these attacks will not have a detrimental impact on the model's performance when evaluated on benign (clean) data \cite{liu2020reflection,zhao2020clean,jiang2022backdoor}.
Our work diverges from these approaches by generating unlearnable examples using imperceptible noise to effectively ``bypass" the training process of DNNs, rendering them incapable of learning from the altered data, thereby offering a more robust strategy for data protection.

\vspace{-1ex}
\subsection{Adversarial Attack}
%Adversarial attacks are a class of techniques aimed at fooling machine learning models, particularly DNNs, by introducing carefully crafted perturbations into the input data. These perturbations are often small and imperceptible to humans but can cause the model to make incorrect predictions or classifications. The primary goal of an adversarial attack is to find the smallest possible alteration to the input that leads to misclassification or a significant increase in prediction error.
Adversarial attacks are techniques designed to deceive machine learning models, especially DNNs, by injecting minor, often imperceptible noise that can lead models to make different predictions. The aim is to identify the minimal input modification causing misclassification or heightened prediction error.
Extensive research has established adversarial examples that can deceive DNNs during the testing phase \cite{szegedy2013intriguing,goodfellow2014explaining,kurakin2016adversarial,carlini2017towards,madry2018towards,croce2020reliable,shafahi2020universal}.
In these attacks, the adversary identifies a form of error-maximizing noise that significantly increases the model's prediction error.
In response to the vulnerabilities exposed by adversarial attacks, adversarial training has emerged as the most robust countermeasure \cite{madry2018towards,zhang2019theoretically,wang2019convergence,wu2020adversarial,wang2019improving,shan2022post}. This training strategy is formulated as a \emph{min-max} optimization problem, where the objective is to minimize the model's vulnerability to error-maximizing noise while maximizing its performance on clean data.
%This dual objective ensures that the model is accurate and resilient against adversarial perturbations.

\vspace{-1ex}
\subsection{Unlearnable Examples}
In contrast to adversarial examples, which focus on error-maximizing noise, unlearnable examples (UEs) pursue the opposite direction by identifying minimal noise that reduces the model's error through a \emph{min-min} optimization process.
In this regard, Huang et al.~\cite{huang2020unlearnable} proposed the concept of UE, aimed at making training data ineffective for DNNs. Similarly, Yuan et al.~\cite{yuan2021neural} introduced Neural Tangent Generalization Attacks (NTGAs), a method that proficiently conducts generalization attacks on DNNs without requiring explicit knowledge about the learning model.
%Remarkably, NTGAs preserve the accuracy of dataset labels, ensuring that legitimate users can still employ the dataset as usual. Moreover, the experimental results highlight the strong transferability of NTGAs across a wide range of models, encompassing fully connected networks and Convolutional Neural Networks (CNNs), even under various training conditions.
Fu et al.~\cite{fu2022robust} identified privacy limitations using error-minimizing noise and introduced robust error-minimizing noise via a \emph{min-min-max} optimization. This limits adversarial learners from gleaning dataset information. Ren et al. \cite{ren2022transferable} introduced transferable UEs that can improve their data-wise transferability. Based on this, Zhang et al. \cite{zhang2023unlearnable} proposed Unlearnable Clusters (UCs), offering a versatile approach to create UEs adaptable to various label exploitations.
On the other hand, several countermeasures have been proposed against unlearnable examples, such as UEraser \cite{qin2023learning} that uses error-maximizing data augmentation, and Jiang et al. \cite{jiang2023unlearnable} propose a method to revert unlearnable samples to learnable ones.
Our work expands the UE from the image domain to the time series domain across classification and generation tasks. Our approach can target the specified segments of data and make them unlearnable, thereby safeguarding the sensitive time series data against misuse and exploitation.

\section{Error-minimizing Noise For Time Series}

In this section, we introduce our proposed method for generating error-minimizing noise \emph{selectively} on segments of time series data.

\vspace{-1ex}
\subsection{Objective}

In this paper, we primarily focus on applications related to time series classification and generation tasks. The models of interest in this domain are Recurrent Neural Networks (RNNs).
The goal of our research is to protect time series samples that contain sensitive information in the public domain from being exploited by RNNs to ensure sensitive details are not inadvertently learned by machine learning models. Consequently, the defender's objectives are twofold. First, given the open accessibility of the data, it is imperative to inhibit deep learning models (RNNs) from processing or learning from this sensitive information. Second, these protective measures should not adversely affect the model's ability to generalize or perform its intended functions using non-sensitive information.

\vspace{-1ex}
\subsection{Threat Model}
The defenders (data subjects) are aware of the general characteristics of the dataset into which their data will be collected and incorporated. This knowledge may include aspects such as the type of data, its source, and its intended application.
While the defenders lack the authority to directly access or modify the dataset, they have the ability to access or alter their own individual data within it. 
Additionally, defenders are aware of the architecture of the DNNs being employed, but they lack information on more granular details such as the exact training procedure, optimization methods, or hyperparameters. This setting simulates the real-world scenario where defenders are often equipped with only partial information and lack full access or a complete understanding of the system.
The defenders seek to safeguard their sensitive information from unauthorized exploitation by introducing error-minimizing noise into the time series data. This addition of noise is designed to render only the sensitive portions of the data unlearnable for machine learning models.
Given the defenders' limited knowledge of the exact training model, the noise introduced should be adaptable across various machine learning models. Defenders, therefore, create noise based on a model they estimate to be close to the real one. This estimated model aids in crafting noise that remains effective across different architectures.

\vspace{-1ex}
\subsection{Challenges}

Building upon the established concept of unlearnable examples in image data \cite{huang2020unlearnable,ren2022transferable,fu2022robust,zhang2023unlearnable}, our research extends this technique to time series data. We aim to create specific unlearnable (error-minimizing) noise that can be added to time series samples, hindering DNNs from effectively learning from these modified samples. While most existing methods focus on images and CNNs, our approach focuses on time series and RNNs. RNNs operate by processing sequential elements and retaining information from prior elements, and this iterative, memory-like nature of RNNs poses unique challenges in generating unlearnable noise. Unlike image data, where noises added to different locations are more independent, noises in time series data can be highly interconnected across the sequence, interfering with each other. Thus, a change at one single time point can cascade effects throughout the sequence. Given the memory mechanism of RNNs, even slight perturbations can amplify in later stages, greatly affecting the final output. Hence, creating error-minimizing noise for RNNs requires a novel approach that ensures the targeted segments of data are unlearnable while preserving the integrity and semantic value of the remaining parts.

\vspace{-1ex}
\subsection{Problem Formulation}

Consider a time series sequence \( x_i \), indexed by time \( t \), which can be formally represented as \( x = \{ \mathbf{x}_{0}, \mathbf{x}_{1}, \ldots, \mathbf{x}_{t-1}, \mathbf{x}_{t} \} \). This sequence is processed through an RNN model for classification task that yields \( y_i = f_{\theta}(x_i) \), where \( y_i \) serves as a class probability vector in the context of time series classification, or as a generated sequence for sequence generation.  \( \theta \) represents the model's learnable parameters, which govern the transformation \( f \). Training the RNN model is to minimize its empirical error on the training samples, which can be achieved via empirical risk minimization (ERM). The optimization problem can be formulated as follows:
\begin{gather}
    \min_{\theta} \E_{(x_i,y_i) \in D}\ell(f_\theta(x_i),y_i).
    \label{eq:erm_rnn_train}
\end{gather}
where $D$ represents the training data and $\ell$ is the loss function that quantifies the dissimilarity between the model's output and the true target. 
% The objective is to find the optimal \( \theta \) that minimizes this loss function across the entire collection of training samples.
To ensure minimal or negligible updates to the model parameters for a given time series sample, we introduce an error-minimizing noise denoted as $\delta$. The primary objective of incorporating this noise is to significantly reduce the training loss of a sample when noise has been added to it.  This noise term is designed to have the same dimensional structure as the input sample $x$, resulting in a sequence of noise values $\delta=\{\delta_{0}, \delta_{1}, ..., \delta_{t-1}, \delta_{t}\}$.
Consequently, when considering the time series sample \( x_i \) in conjunction with its sample-specific error-minimizing noise \( \delta_i \), the combined effect can be mathematically expressed as follows:
\begin{gather}
    \ell(f_\theta(x_i + \delta_i), y_i) \rightarrow 0.
    \label{eq:erm_noise_cause}
\end{gather}
The objective of this perturbation is to drive the loss \( \ell(\cdot) \) towards zero. By doing so, the noise serves to minimize the discrepancy between the RNN's output and the actual target \( y_i \). Consequently, the model is tricked into learning nothing from these perturbed samples. 

\vspace{-1ex}
\subsection{A Straightforward Baseline Approach}

A baseline method can be established for the generation of unlearnable examples in time series data by leveraging the concept of unlearnable examples as described in \cite{huang2020unlearnable}. During the training phase of a basic noise generator, denoted as \( f'_\theta \), the system aims to solve the optimization problem as stated in Equation~\ref{eq:ue_train}.
%The defensive perturbation radius, represented as $\rho_u$, ensures that the error-minimizing noise generated remains unnoticeable, hence preserving the integrity and usefulness of the data while simultaneously protecting it.

\begin{gather}
    \min_{\theta} \frac{1}{n} \sum_{i=1}^n \min_{\|\delta_i\|\leq \rho_u} \ell(f'_\theta(x_i+\delta_i),y_i).
    \label{eq:ue_train}
\end{gather}

The generation of an unlearnable example, represented as \( (x', y) \), is accomplished using the trained noise generator \( f'_\theta \). This transformed data point is formally defined in \Eqref{eq:unlearnable_sample}.

\begin{gather}
x' = x + \arg\min_{\|\delta\|\leq \rho_u} \ell(f'_\theta(x+\delta),y).
\label{eq:unlearnable_sample}
\end{gather}

Given the sequential nature of RNNs, Backpropagation Through Time (BPTT) will be used where the network will be unrolled to match the length of the time series data. The calculation of the loss with respect to this unrolled RNN model takes into account these hidden states, allowing for a more detailed understanding of how each temporal data point in the sequence influences the overall loss.

\subsection{Controllable Noise on Partial Time Series Samples}

A significant limitation of directly translating image-based methods to time series data is the inability to localize and control the region of noise application. In this case, noise tends to be distributed uniformly across the entire sequence.
In the context of fixed-sized inputs, such as images, this uniform distribution is generally acceptable because the noise can be easily processed and interpreted within a consistent framework.
However, given that RNN models process time series data in sequential order across the time regions, the effectiveness of this noise is not uniform across different temporal segments. Consequently, some portions of the time series will be more affected than others, leading to inconsistent training and prediction outcomes.

\begin{figure}[ht!]
    \centering
    \includegraphics[width=\linewidth]{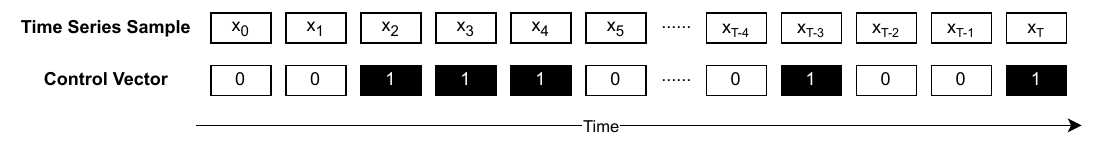}
    \caption{Illustration of the control vector applied on a time series sample of length $T$. Data protection is indicated when the control vector highlights particular time stamps with a value of 1 (marked in black).}
    \label{fig:control_vector}
\end{figure}

We propose a novel control vector, denoted as $v$, that highlights regions within the samples that should be ``protected'' from data exploitation. This concept is depicted in Figure \ref{fig:control_vector}. As an example, consider a dialogue that comprises speech data from multiple individuals. If there is a need to protect the speech of a specific individual, their corresponding temporal segments can be distinctly marked using the control vector.
To achieve this, we selectively add error-minimizing noise to the targeted segments of the time series data. Our primary objective is to reduce the training loss associated with these specified regions of the time series samples by solving the following optimization problem:
\begin{align}
    \min_{\theta} \frac{1}{n} \sum_{i=1}^n \min_{\|\delta_i\|\leq \rho_u} |&\ell(f'_\theta(x_i+\delta_i \odot v_i),y_i) - \notag\\
    \alpha\cdot&\ell(f'_\theta(x_i \odot (1-v_i)),y_i)|,
    \label{eq:cun_train}
\end{align}
where $\odot$ represents element-wise multiplication on two vectors, and $|x|$ represents the absolute value of $x$. Specifically, our objective is to ensure that the training loss incurred by the sample with noise added to the targeted region (\(x_i+\delta_i \odot v_i\)) is equivalent to the training loss when the target region is completely omitted from the time series sample (\(x_i \odot (1-v_i)\)). This is achieved by minimizing the absolute difference between these two loss terms.
By aligning the loss from noise addition to that of complete removal on the partial time series sample, we ensure that the model does not derive any insights from the target regions of the sample, while preserving the consistent learning patterns from the other segments. 
In summary, our method endeavors to provide a new solution that bridges the gap between conventional error-minimizing noise generation methods and the unique requirements of time series data.

\section{Experiments}

In this section, we evaluate our proposed controllable error-minimizing noise in both time series classification and sequence generation tasks.  

\vspace{-1ex}
\subsection{Experiment Setup}

For our experiments on time series classification tasks, we use a simple RNN architecture as the backbone model. This architecture consists of an input layer, the dimensions of which are determined by the feature set of the dataset. The model includes three recurrent hidden layers, each having 64 hidden units, and one output layer.
For training, we adopted a batch size of 256 and used the Adam optimizer with a starting learning rate of 0.01. Specific parameters for RNN training included a 0.01 learning rate for noise generation ($\gamma$), a maximum noise magnitude set to $0.05\times max_\text{magnitude}$ per sample ($\rho_u$), a trade-off parameter of 1 ($\alpha$), a warm-start duration of 5 epochs ($T_\text{warm\_start}$), and a total training epoch of 50 ($T_\text{training}$). We use the model checkpoint at the $55^{th}$ epoch as the final error-minimizing noise generator. Subsequently, we applied three different noise configurations with the control vector \( v \), covering 20\%, 50\%, and 100\% of the sample with 10\% non-overlapping consecutive segments. The positioning of these segments is selected randomly for every sample.

We use ten unique time series datasets, including six univariate datasets from the UCR Archive and four multivariate datasets from the MTS Archive. We also employ two baseline methods including masking and universal adversarial perturbation (UAP) ~\cite{li2020universal}. In our approach, we use masking to hide specific segments within time series samples. We randomly choose segments covering 50\% of each sample, dividing them into five non-overlapping regions, each spanning 10\% of the sample. This masking serves as a baseline to gauge the model's performance with significant data absence. 
To ensure equitable comparison, the adversarial perturbation was capped at $0.05\times max_\text{magnitude}$ and integrated into 50\% of every sample, specifically at the same regions chosen for masking.

\vspace{-1ex}
\subsection{Against Classification Models}

\begin{table}[t]
\setlength{\tabcolsep}{0.6em}
\renewcommand{\arraystretch}{1.5}
\centering
\caption{Performance degradation results of various noise types introduced into the training data. Datasets $D_{1}$ through $D_{6}$ are univariate, sourced from the UCR Archive; whereas $D_{7}$ to $D_{10}$ are multivariate from the MTS Archive. The $2^{nd}$ column, labeled as \textbf{Clean}, depicts the accuracy of models trained on benign data.}
\scalebox{0.7}{
\begin{tabular}{lcccccccccccc}
\toprule
\multicolumn{1}{l}{}  \makecell{\multirow{1}{*}{\textbf{Dataset}}}   &  \textbf{Clean}   & \textbf{Masking}      & \textbf{Universal}             & \textbf{Ours$_{(20\%)}$}                    & \textbf{\textbf{Ours$_{(50\%)}$}}    & \textbf{\textbf{Ours$_{(100\%)}$}}   \\ \midrule
($D_{1}$) BirdChicken  &  96.0\%  &  80.9\% (-15.1\%)  &  39.1\% (-56.9\%)  &  19.3\% (-76.7\%)  &  12.1\% (-83.9\%)  &  8.8\% (-87.2\%)  \\
($D_{2}$) ECG5000  &  94.6\%  &  78.4\% (-16.2\%)  &  41.6\% (-53.0\%)  &  16.9\% (-77.7\%)  &  11.5\% (-83.1\%)  &  7.4\% (-87.2\%)  \\
($D_{3}$) Earthquakes  &  72.5\%  &  65.1\% (-7.4\%)  &  27.7\% (-44.8\%)  &  10.7\% (-61.8\%)  &  6.4\% (-66.1\%)  &  3.1\% (-69.4\%)  \\
($D_{4}$) ElectricDevices  &  72.3\%  &  63.3\% (-9.0\%)  &  22.7\% (-49.6\%)  &  10.5\% (-61.8\%)  &  7.6\% (-64.7\%)  &  5.0\% (-67.3\%)  \\
($D_{5}$) Haptics  &  50.2\%  &  29.0\% (-21.2\%)  &  17.5\% (-32.7\%)  &  13.4\% (-36.8\%)  &  8.3\% (-41.9\%)  &  6.3\% (-43.9\%)  \\
($D_{6}$) PowerCons  &  88.2\%  &  68.4\% (-19.8\%)  &  37.0\% (-51.2\%)  &  15.6\% (-72.6\%)  &  9.3\% (-78.9\%)  &  4.9\% (-83.3\%)  \\ %\midrule
($D_{7}$) ArabicDigits  &  99.4\%  &  83.0\% (-16.4\%)  &  26.1\% (-73.3\%)  &  9.4\% (-90.0\%)  &  4.7\% (-94.7\%)  &  2.1\% (-97.3\%)  \\
($D_{8}$) ECG  &  87.4\%  &  74.2\% (-13.2\%)  &  20.2\% (-67.2\%)  &  8.9\% (-78.5\%)  &  5.6\% (-81.8\%)  &  2.6\% (-84.8\%)  \\
($D_{9}$) NetFlow  &  89.4\%  &  78.5\% (-10.9\%)  &  16.7\% (-72.7\%)  &  6.7\% (-82.7\%)  &  3.2\% (-86.2\%)  &  1.9\% (-87.5\%)  \\
($D_{10}$) UWave  &  93.4\%  &  80.8\% (-12.6\%)  &  26.4\% (-67.0\%)  &  10.2\% (-83.2\%)  &  7.9\% (-85.5\%)  &  4.0\% (-89.4\%)  \\

\bottomrule
\end{tabular}
}
\label{table:exp_CUN_TSC}
\end{table}

Our experimental results for the controllable unlearnable noise generator, featuring three configurations and two baseline methods, are presented in Table \ref{table:exp_CUN_TSC}. Using the masking baseline, we noticed a 14.18\% average drop in accuracy compared to the clean model. The unmasked segments, retaining key features, possibly account for the limited decline. With the time series UAP, the accuracy decrease averaged 56.84\%. Remarkably, our proposed noise method, targeting only 20\% of samples, achieved a more significant average accuracy drop of 72.18\%, emphasizing its efficacy over the UAP. With 50\% targeting, as in the baselines, accuracy fell to 7.66\%, marking a 76.68\% reduction from the clean model. Additionally, our error-minimizing noise demonstrates more significant impacts on multivariate datasets, which capture the interactions and relationships across multiple variables. This multi-dimensionality allows the unlearnable noise to envelop both primary and subtle features. As a result, it can reduce the training loss more effectively, obscuring the genuine data patterns. 

Taking a closer look at the accuracy drops, we found that increasing the amount of unlearnable noise does not linearly decrease classification accuracy. This suggests a diminishing return on increasing noise levels, indicating that beyond a certain threshold, the addition of more unlearnable noise might not yield significantly enhanced privacy protections. This observation implies that introducing noise to only a segment of the time series might be the most beneficial strategy. By targeting only a small part of the samples, one can achieve the desired reduction in training effectiveness without compromising the entire dataset. The results highlight the efficacy of our method, showcasing its adaptability in safeguarding time series data privacy, especially potent against data misuse in classification tasks.

\vspace{-1ex}
\subsection{Against Generative Models}

We extend the evaluation to assessing the application of our proposed unlearnable noise in the context of time series generation tasks. We employ 8 multivariate time series datasets for this study, encompassing a range of classes and sample sizes. For the task of data generation, we apply two time series generative models: the Recurrent GAN (RGAN) \cite{esteban2017real} and Quant GAN (QGAN) \cite{wiese2020quant}. These models are then used to generate synthetic data for the first class (class 0) of each dataset. We follow the training procedure stated in the original papers. The noise is configured to perturb 50\% of the samples in the target class, and every selected sample is entirely perturbed by the noise.

We apply the \emph{Train on Synthetic, Test on Real} (TSTR) \cite{esteban2017real} approach to test the effectiveness of our proposed unlearnable noise. Specifically, we first train a GAN model with data perturbed by unlearnable noise, then train a classifier model using data generated by the GAN and subsequently test it on a separate set of genuine samples. In this experiment, we subset all samples from the first class (class 0) of each dataset and then feed them for GAN training. The objective is to minimize the generator's reconstruction loss on the entire sample. Then, we train the time series classifiers using the generated synthetic samples, using Long Short-Term Memory (LSTM) and Fully Convolutional Network (FCN).

\begin{table}[h!]
\setlength{\tabcolsep}{0.6em}
\renewcommand{\arraystretch}{1.08}
\centering
\caption{Classification accuracy of real or synthetic time series samples using the "Train on Synthetic, Test on Real" (TSTR) approach. The $2^{nd}$ column, labeled as \textbf{Real}, depicts the accuracy of classification models trained and tested on benign data. The columns presented as \textbf{Model$_{c}$} use clean data to train the generative model. The columns presented as \textbf{Model$_{n}$} use unlearnable data to train the generative model. }
\scalebox{0.95}{
\begin{tabular}{llr|rr|rr}
\toprule
\textbf{Dataset}                         & \textbf{Network} & \multicolumn{1}{l}{\textbf{Real}} & \multicolumn{1}{|l}{\textbf{RGAN$_{c}$}} & \multicolumn{1}{l|}{\textbf{RGAN$_{n}$}} & \multicolumn{1}{l}{\textbf{QGAN$_{c}$}} & \multicolumn{1}{l}{\textbf{QGAN$_{n}$}} \\ \midrule
\multirow{2}{*}{($D_{7}$)}  & FCN     & 99.6\%                   & 75.2\%                          & 6.2\%                           & 78.6\%                          & 8.4\%                           \\
                                         & LSTM    & 98.4\%                   & 83.4\%                          & 4.2\%                           & 81.4\%                          & 2.6\%                           \\
\multirow{2}{*}{($D_{8}$)}           & FCN     & 91.2\%                   & 77.6\%                          & 7.4\%                           & 76.0\%                          & 7.8\%                           \\
                                         & LSTM    & 89.4\%                   & 72.0\%                          & 3.0\%                           & 73.6\%                          & 5.8\%                           \\
\multirow{2}{*}{($D_{9}$)}       & FCN     & 94.6\%                   & 75.4\%                          & 11.6\%                          & 77.0\%                          & 10.6\%                          \\
                                         & LSTM    & 90.1\%                   & 74.0\%                          & 6.8\%                           & 75.6\%                          & 3.4\%                           \\
\multirow{2}{*}{($D_{10}$)}        & FCN     & 95.0\%                   & 82.0\%                          & 10.5\%                          & 84.0\%                          & 13.6\%                          \\
                                         & LSTM    & 93.0\%                   & 86.0\%                          & 5.2\%                           & 88.0\%                          & 6.2\%                           \\
\multirow{2}{*}{($D_{11}$)}       & FCN     & 94.2\%                   & 80.4\%                          & 8.4\%                           & 83.2\%                          & 9.6\%                           \\
                                         & LSTM    & 95.0\%                   & 76.4\%                          & 3.8\%                           & 82.4\%                          & 5.8\%                           \\
\multirow{2}{*}{($D_{12}$)} & FCN     & 78.9\%                   & 54.2\%                          & 6.2\%                           & 58.0\%                          & 8.4\%                           \\
                                         & LSTM    & 76.0\%                   & 57.8\%                          & 2.8\%                           & 60.8\%                          & 4.0\%                           \\
\multirow{2}{*}{($D_{13}$)}       & FCN     & 86.4\%                   & 68.6\%                          & 11.6\%                          & 73.6\%                          & 10.8\%                          \\
                                         & LSTM    & 75.0\%                   & 64.2\%                          & 6.0\%                           & 72.4\%                          & 5.2\%                           \\
\multirow{2}{*}{($D_{14}$)}      & FCN     & 71.0\%                   & 54.6\%                          & 9.5\%                           & 61.2\%                          & 10.2\%                          \\
                                         & LSTM    & 64.0\%                   & 49.0\%                          & 5.4\%                           & 56.0\%                          & 4.8\%                          \\
\bottomrule
\end{tabular}
}
\vspace{-1ex}
\label{table:exp_CUN_Gen}
\end{table}

The experimental results shown in Table \ref{table:exp_CUN_Gen} demonstrate a significant drop in performance when adding unlearnable noise to 50\% of the training samples. The average classification accuracy drops below 10\%, marking an average reduction of over 60\% when compared to the results obtained for clean data training. 
Note that, while the noise is introduced into only 50\% of the samples within a specific class, it has the capability to render the entire class unlearnable (non-generative) against the sequence generation model.
This implies that our proposed noise has great potential to be applied to protect sensitive samples from being learned during the training of a generative model, preventing the model from recreating or understanding the sensitive or private aspects of the original data.

\section{Conclusion}

In this work, we have studied the problem of protecting time series data against unauthorized exploitations. We extended the concept of Unlearnable Examples (UEs) from the image domain to the time series domain and proposed a novel method specifically designed for generating unlearnable noise for time series. The proposed method leverages a novel min-min bilevel optimization framework alongside a control vector, enabling the creation of unlearnable noise targeted at the most sensitive parts of a time series. This approach can be selectively used on specific segments of the time series data. Through extensive experiments on both time series classification and generation tasks, we demonstrated the effectiveness of our method across different datasets. Our work could help individuals and organizations protect their time series data from being exploited (without permission) in the development of commercial models.

\bibliographystyle{splncs04}
\bibliography{ref}

\end{document}